\documentclass{article}
\usepackage{iclr2021/iclr2021_conference,times}

\usepackage{amsmath,amsfonts,bm}

\def\eqref#1{equation~\ref{#1}}

\def\1{\bm{1}}

\DeclareMathAlphabet{\mathsfit}{\encodingdefault}{\sfdefault}{m}{sl}
\SetMathAlphabet{\mathsfit}{bold}{\encodingdefault}{\sfdefault}{bx}{n}

\DeclareMathOperator*{\argmin}{arg\,min}

\iclrfinalcopy

\usepackage[utf8]{inputenc} 
\usepackage[T1]{fontenc}    
\usepackage{hyperref}       
\usepackage{url}            
\usepackage{booktabs}       
\usepackage{amsfonts}       

\usepackage{amsmath}
\usepackage{amssymb}
\usepackage{graphicx}
\usepackage{subcaption}
\usepackage{tabularx}
\usepackage{floatrow}
\usepackage{algorithm}
\usepackage[noend]{algpseudocode}
\usepackage{wrapfig}

\usepackage{multirow}

\date{}

\title{Evaluating Online Continual Learning \\with CALM}

\author{%
  Germán Kruszewski\thanks{Work done while the authors were at Facebook AI.}\\
  Naver Labs Europe \\
  \texttt{german.kruszewski@naverlabs.com}
  \And
  Ionut Teodor Sorodoc\\
  Pompeu Fabra University\\
  \texttt{ionut.sorodoc@gmail.com} \\
  \And
  Tomas Mikolov$^{*}$\\
  CIIRC CTU\\
  \texttt{tmikolov@gmail.com}\\
}
\begin{document}
\maketitle
\begin{abstract}
    Online Continual Learning (OCL) studies learning over a continuous
    data stream without observing any single example more than once, a setting
    that is closer to the experience of humans and systems that must learn
    ``on-the-wild''.
    Yet, commonly available benchmarks are far from these real world
    conditions, because they explicitly signal different tasks, lack latent
    similarity structure or assume temporal independence between different
    examples.
    Here, we propose a new benchmark for OCL based on language modelling in
    which input alternates between different languages and domains without any
    explicit delimitation.
    Additionally, we propose new metrics to study catastrophic forgetting in
    this setting and evaluate multiple baseline models based on compositions of
    experts.
    Finally, we introduce a simple gating technique that learns the latent
    similarities between different inputs, improving the performance of a
    Products of Experts model.
\end{abstract}

\section{Introduction}
\label{sec:introduction}

Machines, like humans, can learn to perform multiple different tasks
from feedback alone~\citep{Caruana:1997}. 
On the other hand, humans, but not machines, can benefit from settings in which
tasks are presented repeatedly for multiple trials before switching to the
next one~\citep{Flesch:etal:2018}, whereas machines require examples to be
presented in a shuffled (\emph{i.i.d}) order to learn effectively.
Otherwise, they suffer from an effect known as ``catastrophic forgetting'' or
``catastrophic interference''~\citep{mccloskey1989catastrophic,
ratcliff1990connectionist}.
While there has been a considerable amount of work focused on solving this
problem, an endeavour that also goes by the name of `Continual', 
`Incremental' or `Life-long' Learning, a large part of it is evaluated on settings in which
there is an explicit delimitation signal for every new task presented to the
model \citep{kirkpatrick2017overcoming, Zenke:etal:2017, Sodhani:etal:2018,
serra2018overcoming,LopezPaz:Ranzato:2017, fernando2017pathnet, lee2017overcoming,
rusu2016progressive, li2018learning, aljundi2017expert,Adel:etal:2020,Titsias:etal:2020,Ebrahimi:etal:2020,vonOswald:etal:2020,Li:etal:2020,Yoon:etal:2020}.
However, humans seem to learn without any such signalling. 
Moreover, the concept of ``task'' could be vacuous, as it could be represented
by shifting data distributions~\citep{Lesort:etal:2020}. 
%

Even though  the emerging field of Online Continual
Learning~\citep{Parisi:Lomonaco:2020,Aljundi:etal:2019b} or Task-Free Continual
Learning \citep{Aljundi:etal:2019,Lee:etal:2020} has started to propose
solutions to these problems, commonly available benchmarks make assumptions
that are far from the real world conditions, such as lacking latent similarity
structure on the data stream (e.g. orthogonal permutations of the pixels in an
image) or assuming temporal independence between different examples (e.g. an
image of
a chair can be classified as ``chair'' independently of any previous examples).
Consider, instead, the challenge of natural language learning which requires
making sense of a highly correlated and temporally interdependent data stream.
We argue that the notable scarcity of benchmarks featuring temporally
correlated sequences of examples, with short and long-term dependencies, latent
similarities between different classes of examples, and no explicit delimitation
when transitioning between different classes has left a blind spot in the
Online Continual Learning community, which we address here.
Moreover, almost none of the commonly used benchmarks deals with language,
further limiting the amount of research that extends to this modality.

Here, we make a two-fold contribution towards studying online continual
learning in neural networks in a linguistic setting.
First, we bring CALM (\emph{\underline{C}lass-\underline{A}gnostic Continual
\underline{L}anguage \underline{M}odelling}) to the community, a continual
language modelling evaluation framework containing text that alternates between
different classes of input (e.g. different languages or domains) with
latent similarities to which the models could adapt.
We introduce two variants.
The first is a character-based language modelling benchmark featuring 
five different languages that randomly switch between one another.
The second one is a word-based language modelling task, where the text
alternates between four different domains.
No segmentation signal is given when a switch happens, thus requiring models
to learn to adapt to these changes.
We also propose novel metrics that capture the impact of
catastrophic forgetting in an online learning setting by measuring how
efficiently can models adapt to class switches.
In line with \cite{Aljundi:etal:2019}, we note that when a distribution
shift occurs, a neural network that suffers from catastrophic 
forgetting will display a spike in the loss signal, even when the distribution
had been observed in the past (see Figure \ref{fig:task}).
Thus, we propose catastrophic forgetting metrics based on characterizing the
size of these peaks.
The benchmark is provided as a Python library that can be easily imported into a
PyTorch project.\footnote{\label{url}Code and materials are available at \url{https://github.com/germank/calm}}
Second, we evaluate multiple baselines based on expert architectures and  propose a
novel albeit simple mechanism that we call \emph{plastic gates}, which we 
show to improve the performance of Products of Experts.
Our post-hoc analysis shows that this mechanism is effective in producing a
gating strategy that helps to circumvent catastrophic interference while also
uncovering latent similarities in the input classes.

\section{Related work}
\label{sec:related-work}

The field of Continual Learning, Incremental Learning or Lifelong Learning 
has grown to encompass a large body of work, which is better summarized in
respective reviews~\citep{Parisi:etal:2019, Lesort:etal:2020}.
An overwhelming majority of this work concerns image classification problems or
object recognition.
Some evaluation datasets are derived from traditional machine learning datasets
by manipulating the input examples in more or less artificial ways --like
Permuted MNIST~\citep{kirkpatrick2017overcoming} or Rotated
MNIST~\citep{LopezPaz:Ranzato:2017}-- while others keep examples unchanged but
present them in a specific non-i.i.d.\ order, like for instance,
iCIFAR-100~\citep{rebuffi2017learning} or split-MNIST~\citep{Zenke:etal:2017}.
All of these datasets comprise single-input classification problems in which
there are no temporal dependencies nor correlations between two consecutive
examples.
To better approximate the conditions of real-world experiences,
\citet{Fanello:etal:2013}, \citet{Lomonaco:Maltoni:2017} and \citet{Roady:etal:2020} introduced
iCubWorld, CORe50, and Stream-51 respectively, which comprise short videos of objects from
different angles (further including naturalistic scenes in the latter case).
These datasets address the problem of correlated examples, but not of
temporal dependencies, which we do address in this work.
\citet{Li:etal:2020} and \citet{dAutume:etal:2019} proposed the only benchmarks
dealing with language that we know of, in which the former adopts a sequence to
sequence paradigm to study incremental learning of new vocabulary items on
simplified or artificial datasets, while the latter adapted existing text
classification and QA benchmarks analogously to above-mentioned work in image
classification.
Our work instead uses naturalistic textual data containing natural latent
similarities between distributions that can drive information transfer or
forgetting.

By and large, work directed to address catastrophic forgetting in neural
networks presumes the existence of a task identifier to signal different
learning units. 
However, recent work has aimed at tackling catastrophic forgetting even in
conditions in which no task boundaries are provided~\citep{Aljundi:etal:2019,
Lee:etal:2020}, going under the name of ``Task-Free Continual
Learning'' or ``Online Continual Learning''~\citep{Parisi:Lomonaco:2020,Aljundi:etal:2019b}.
Of these works, only \hbox{\citet{Aljundi:etal:2019}} uses naturalistic data to
classify actors appearing in soap-opera episodes~\citep{Aljundi:etal:2016},
while others resort to artificially modified datasets like split or permuted
MNIST.
Here, we complement this resource with a text-based benchmark for Task-Free
Continual Learning, while arguing for more work on more naturalistic non-i.i.d.
datasets.

Another aspect of Continual Learning deals with how models are evaluated.
Most often, this is done by measuring accuracy on a dedicated test set~\citep{
LopezPaz:Ranzato:2017,Diaz:etal:2018,Hayes:etal:2018,Chaudhry:etal:2018,
dAutume:etal:2019}.
However this evaluation protocol is tailored for batch learning conditions, 
in which a model is fit to a training dataset, and then stops learning. 
Here, instead, we argue in favour of situated evaluation protocols adapted to
far-from-equilibrium learning conditions~\citep{Holland:1992} by adopting an
Online Learning framework~\citep{Hoi:etal:2018}, which is also known as the prequential
approach~\citep{Dawid:1984,Gama:etal:2013}.

On the modelling side, this work explores Mixture of
Experts~\citep{jacobs1991adaptive} and Product of
Experts~\citep{hinton1999products} architectures. 
Variations thereof are at the base of many architectural proposals for
addressing catastrophic
forgetting~\citep{rusu2016progressive,li2018learning,aljundi2017expert,Lee:etal:2020}.
However, often they are accompanied by other mechanisms, such as the growth of
new modules, freezing of weights or generative modelling of the input.
Here we examine the simplest enactments of these architectures and propose
an easy-to-implement gating mechanism which can be learned online and 
provides a strong baseline for more complex architectures.

Finally, our study falls within the line of language modelling using
 neural network models \citep{bengio2003neural,mikolov2010recurrent}.
In this context, adaptation to the recent past has been studied in the context
of cache models \citep{grave2017unbounded,merity2016pointer}.
There, a non-parametric model deals with capturing high-frequency recent
statistics while a parametric model captures the more stable aspects of the
distribution. These solutions, however, are not well-adapted for cases in which
the whole distribution changes over time.
Moreover, language modelling is generally studied in a train-test split, where
a model is fitted to the training data and asked to generalize over the unseen
test data. Here, instead, we study how a model can adapt to incoming
data in an online fashion.

\section{The CALM benchmark}
\label{sec:task}
We designed a benchmark for evaluating Online Continual Learning algorithms
having in mind the following three desiderata: 1) naturally correlated
sequential data, 2) task agnosticism and 3) temporally situated evaluation.
\citet{Parisi:Lomonaco:2020} discusses the first two.
The first requires that on the one hand, data is observed in a potentially
infinite data stream with high-dimensional, non-stationary, and temporally
correlated examples.
The second, that learning systems should not be
fed external task boundaries to help them learn in these conditions. 
Furthermore, we also introduce a third desideratum, by which we ask models to
be evaluated \emph{in-situ} on each example presented to the model, following
the classical Online Learning
setting~\citep{Hoi:etal:2018,Sahoo:etal:2018}.
We thus propose an Online Continual Learning benchmark featuring a language
modelling objective where the data stream can switch between different
distributions.
Because switches are not announced to the model, this is a
``Single-Incremental-Task`` or ``No task label'' scenario under the framework
proposed by \citet{Lesort:etal:2020}.

More precisely, consider a sequence of observations $x_t \in \mathcal{X}$
that are fed to a model $M$ parametrized by $\Theta_t$, which makes the
prediction $\hat{y}_t \in \mathcal{Y}$. 
Then, the true target $y_t \in \mathcal{Y}$ will be revealed and the loss $L_t
= L(\hat{y}_t, y_t)$ is observed and later used to compute the model's
performance from a given time $S$ until time $T$ as the average loss in that span
$\bar{L}_S^T = \frac{1}{T-S}\sum_{t=S}^T{L_t}$ for evaluation purposes.
Only after reporting the loss can the model be trained based on the received
feedback, preventing data leakage.
In practice, these examples are presented as mini-batches $(X_t, Y_t) \in
\mathcal{X}^{b\times{}w} \times \mathcal{Y}^{b\times{}w}$ containing $b$
parallel streams, and chunked into small windows of length $w$ for efficiency
considerations related to the training of neural
networks~\citep{Parisi:Lomonaco:2020}.

The data stream is composed of $N$
sequences of consecutive mini-batches of length $T_1, T_2, \dots, T_N$, and starting
at positions $S_i = \sum_{j=1}^{i-1}{T_j}$. 
In turn, each of these sequences belong to one of $n$ different classes
$[\mathcal{D}_1, \dots, \mathcal{D}_n]$, presented in random order.


To characterize the effect of forgetting we note that a model that becomes
disadapted to a given distribution will display a spike in the
loss after the stream switches to this distribution, even if it has been
observed before (see Figure \ref{fig:task}).
For a model to be resilient to catastrophic forgetting, it must adapt quickly
to every new distribution, which can be characterized by the height and width
of these peaks. 
We thus propose the following metrics to complement the standard online performance:

\begin{itemize}

\item \textbf{Loss after switch}: Tracks the loss for the first $k$ times-steps
    after a switch occurs to quantify the height of the peak. Formally,
    $\operatorname{L@sw} = \frac{1}{N}\sum_{i=1}^N{\bar{L}_{S_i}^{S_i+k}}$

    \item \textbf{Recovery time after switch}: Counts the number of time-steps that it
    takes the model to reach the mean loss observed for the last seen sequence of
    the current class. In this way, we can quantify the length of the peak.
\end{itemize}

\subsection{Dataset}
\label{sec:dataset}

In this work, we created two datasets for CALM. 
One is character-level and multilingual, whereas the other is word-level and
multi-domain.
Both benchmarks feature conflicting learning signals when moving between
domains or languages, making the learning systems susceptible to catastrophic 
forgetting.

For our first dataset (\textbf{MultiLingual} and character-based), we propose a
language modelling benchmark in which incoming text data can alternate between
different languages.
This benchmark is character-based because there would hardly be any forgetting
at the word level, as the word distributions hardly share any support.
Concretely, we build on parts of the news corpus developed for the 2009
Workshop of Machine Translation \citep{Callison-Burch2009}.
We extracted text from five languages: English, French, Spanish, German, and
Czech (containing 1.8B, 572M, 160M, 715M and 439M characters,
respectively) because they all have similar character sets, while also showing
interesting linguistic variability.
In particular, they belong to three different Indo-European branches: Romance
(French and Spanish), Germanic (English and German), and Slavic (Czech).
As a consequence, there is a latent similarity structure between the different
classes that models could learn to recognize.
Compared to earlier multilingual corpora~\citep{kawakami2017learning}, our
dataset was carefully constructed to include only linguistically valid 
characters, in order to prevent non-linguistic noise from interfering with
our experiments.
For this, we removed all lines from the input that contained characters
appearing less than 100 times on the full corpus.
The resulting character vocabulary consists of 211 characters.
 
The second dataset is an English word-level \textbf{MultiDomain} dataset. 
For this, we used four different source corpora: news (same as above), europarl
\citep{koehn2005europarl}, the British National Corpus \citep{bnc2007british},
and Wikipedia \citep{merity2016pointer}. They each have 300M, 54M, 100M and 101.4M
tokens, respectively. 
In contrast with the previous dataset, word-level is the most appropriate
choice here, as differences between the distributions at the character
level would be too nuanced to drive any forgetting.
We kept in the vocabulary the top 25K words for each corpus, which after
merging yielded a vocabulary size of 58K words.
Samples from all source corpora are included in the appendix.

We then created the final MultiLingual and MultiDomain corpora by joining
$N=100$ different fragments evenly distributed among the different classes
(languages or domains) with lengths sampled from a (truncated) exponential
distribution: $T_i \sim Exp(\lambda)$.
Thanks to this distribution's memorylessness property, it is virtually
impossible to estimate when the next switch is going to happen.
While we do not constrain switches to occur at word or sentence boundaries, but
rather after an integer number of sequences of length $w$, the noise introduced
at transition points for this reason is relatively mild and does not affect
the distribution-alternating nature of the dataset.
In benefit, training and further analysis become considerably simplified by
removing the need to handle variable-length input.
We constructed two different variations with shorter or longer fragments.
For MultiLingual, we constructed 1M and 10M-characters-long corpora
with expected fragment lengths of $\lambda=10$k and $\lambda=100$k characters,
respectively.
For MultiDomain we followed the same procedure, extracting 100
alternating sequences with mean lengths of $\lambda=10$k and $\lambda=20$k, 
for a total of 1M and 2M words. 
These relatively modest sizes allow for faster iteration and exploration of 
different models, while still allowing us to observe forgetting (or lack
thereof) dynamics in the studied models.
To facilitate further research, we release a Python library
providing a data iterator for both datasets in
which a researcher can experiment with different variations by picking
parameters $N$ and $\lambda$.

\section{Baseline Models}
\label{sec:model}

To endow CALM with simple and yet, strong baselines, we explored architectures
based on
(Weighted) Product of Experts or PoE ~\citep{hinton1999products} and Mixture of
Experts or MoE ~\citep{jacobs1991adaptive,eigen2013learning}, henceforth 
generically denoted expert architectures. 
Thanks to combining predictions from different experts, they can potentially
learn different parts of the latent distributions.
Moreover, gating weights can avert catastrophic forgetting on the individual
experts by modulating the learning signal, making them an excellent candidate to
model Online Continual Learning problems. 
Indeed, while variations thereof have been explored before (see
Section~\ref{sec:related-work}), here we emphasize simplicity as it would
befit baseline models, yet not neglecting performance.

In the standard implementation of expert architectures, mixture weights are
produced by a third ``gating'' module as a function of the current inputs.
While this gating model could quickly adapt to changes in the environment,
learning to do so is far from trivial in a continual learning setup, sometimes
requiring pre-training to distinguish input classes~\citep{aljundi2017expert}.
The problem comes from the fact that the gating network must learn a latent
classifier to pick the experts best adapted to the current class, but classes
are observed non-i.i.d. as long sequences of examples from one class at a time.
Thus, the gating network can easily settle for a constant function for any
given current class, which only changes when examples of a different class
start to be observed, making experts vulnerable to catastrophic forgetting.
In order to alleviate this issue and make experts more stable, we propose
\emph{plastic gates}, by which the gates are fast-adapting parameter
values that are trained on recent experience.


More formally, an expert architecture is composed of a set of modules
$\mathcal{M} = \{M_1, \dots, M_n\}$ with parameters $\Theta_{M_1}, \dots,
\Theta_{M_n}$, used to compute a unique prediction as follows.
When an input $x$ (with target $y$) is observed, it is fed to all modules
$M_{1\dots{}n}$, obtaining log-linear outputs
$\mathbf{\tilde{y}}^{(1)}=M_1(x), \dots, \mathbf{\tilde{y}}^{(n)}=M_n(x)$. 
Then, an additional vector of mixture weights $\mathbf{w} \in \mathbb{R}^n$ is used
to combine them.
This vector is computed by a separate gating module $\mathbf{w} =
G(x)$ with parameters $\Theta_G$, jointly trained with the rest of the network.
The output of the full model $\mathbf{y}$ is then a linear combination of
the individual modules outputs $\mathbf{\tilde{Y}}= [\mathbf{\tilde{y}}^{(1)}, \dots, \mathbf{\tilde{y}}^{(n)}]$ weighted by $\mathbf{w}$\footnote{
Note that the since $\tilde{y}^{\text{PoE}}$ linearly combines the logits is is effectively computing a geometric combination of each individual
module's unnormalized probabilities: $\exp(\mathbf{\tilde{y}}^{\text{PoE}}_j)
\propto \prod_{i=1}^n{\exp({\mathbf{\tilde{y}}^{(j)}_i})^{w_i}}$. 
}, after or before normalizing, depending on whether the model is MoE or PoE:
\begin{align*}
    \mathbf{\tilde{y}}^{\text{MoE}}(\mathbf{w})  =  \sum_{i=1}^n{\operatorname{softmax}(\mathbf{w})_i\left(
\operatorname{softmax}\mathbf{\tilde{y}}^{(i)}\right)}    & &\mathbf{\tilde{y}}^{\text{PoE}}(\mathbf{w}) =  \operatorname{softmax}\left(\mathbf{\tilde{Y}^\intercal}\mathbf{w}\right)
\end{align*}
Note that in contrast to MoE, PoE are more efficient to compute because they do
not require to normalize the output of each individual model.
Once the loss is computed on a mini-batch $(X_t, Y_t)$ and kept for evaluation
(see Section \ref{sec:task}), all sub-networks $G$ and $\mathcal{M}$ are
trained for one or more gradient steps to reduce this loss, and the system
moves to the next mini-batch.


\textbf{Plastic Gates} Rather than learning a gating network, which can
be challenging, we propose 
to continually learn the gating coefficients that best fit the recent experience:
\begin{align*}
    \mathbf{w_{t+1}} = \argmin_{\mathbf{w}}{L(\tilde{Y}_t(\mathbf{w}), Y_t)}
\end{align*}
In practice, we perform a (hyperparameter) number $k$ of gradient descent steps 
on the above objective to allow for some regularization of the gates over time.

\paragraph{Parametrization for Language Modelling}
\label{sec:model-param}
We instantiate the expert modules $M_i$ to be double-layered LSTM networks~\citep{hochreiter1997long}, with predictions $\mathbf{\tilde{y}_t}^{(i)}, \mathbf{h_{t+1}}^{(i)} = \textrm{LSTM}_i(x_t, \mathbf{h_t}^{(i)})$.
For the regular gating network, we use a single-layer LSTM network.
That is, $\mathbf{w_t}, \mathbf{h'_{t+1}} = \mathrm{LSTM}(x_t, \mathbf{h'_t})$. 

\section{Experiments}
\label{sec:experiments}


We explored the performance of different baseline models while they made a single
pass over the CALM datasets.
Following standard practice, rather than reporting the cross-entropy loss, we
use the perplexity at each time step, given by $exp(L_t)$.
Furthermore, we allowed the models to learn over the first half of the datasets
without being evaluated, and only start computing metrics on the second half.
Otherwise, we use the measures discussed in Section \ref{sec:task} to track
models' performance, namely, average perplexity (\textbf{ppl}), average
perplexity for $k=10$ batches after a switch (\textbf{ppl@sw})  and recovery
time after a switch (\textbf{rec}).

We explored models featuring different degrees of modularization, varying their
hidden size vectors to make them all have an approximately equal total number
of parameters.
On one extreme, we had a large two-layers \textbf{LSTM} network.
Next, we considered standard \textbf{PoE} and \textbf{MoE} models with mixture
weights computed by an LSTM gating network, plus their plastic weights variants
(\textbf{+PW}), as described in Section \ref{sec:model-param}. 
Moreover, we trained ensemble models (\textbf{Ensemble}), which are equivalent
to a MoE where all mixture weights are $\frac{1}{n}$ for all $n$ modules.
We studied both a more centralized network composed of 5 modules and larger
hidden dimensionality (marked with \textbf{5}) and a more distributed network
with 30 modules but with smaller hidden sizes (marked with \textbf{30}).
As reference points (but not as real contenders), we also trained independent LSTMs (\textbf{Ind. LSTM}),
one for each class, which enabled us to compare the performance of our model
to a situation where there is no forgetting from conflicting learning signals,
but also where there is no possibility of transferring learned representations
across possibly related domains. 
Furthermore, we compare a Mixture-of-Softmax (\textbf{MoS})
model~\citep{Yang:etal:2018}, in which multiple softmax layers are combined to
extract the predictions from a single LSTM module.
While we were also interested in applying state-of-the-art online continual
learning methods~\citep{Lee:etal:2020,Aljundi:etal:2019b}, having these systems
being designed for image classification datasets they would require
non-trivial adaptations significantly departing from the original
models, which would limit any possible conclusions we could draw.
Similarly, we experimented extensively on validation data with Transformer
models~\citep{vaswani2017attention}.
However, due to these models sensitivity to
dataset size and learning rates scheduling schemes which have been studied
extensively for batch-learning~\citep{Papel:etal:2018}, but not for these
far-from-equilibrium~\citep{Holland:1992} conditions, their performance was
worse than expected. 
We give a detailed account of our attempts in the appendix and
leave a study on how to adapt these models for Online Continual Learning for
future work.

We controlled the number of model parameters to remain constant for each of the
MultiLingual (about 21M parameters) and the MultiDomain (about 600M parameters)
experimental setups.
(The difference in size is explained by the larger vocabulary sizes in the
latter.)
For this, we adjusted the hidden dimensionality of different models
accordingly, which, together with all explored hyperparameters, are reported in
the appendix.
We kept the size of the incoming batches fixed at $w=20$ and $b=10$ for all
models and used PyTorch \citep{paszke2017automatic} with the standard
implementations for the underlying models.

\subsection{Results}
\label{sec:results}
\begin{table*}[ht]
    \setlength{\tabcolsep}{3.5pt}
    \setlength{\aboverulesep}{0pt}
    \setlength{\belowrulesep}{0pt}
    \renewcommand{\arraystretch}{1.2}
        \begin{small}
        \centering
        \begin{tabular}{ccccccccccccc}
        \cline{2-13}
         & \multicolumn{6}{c}{MultiLingual} & \multicolumn{6}{c}{MultiDomain}\\         
         \cmidrule(lr){2-7} \cmidrule(lr){8-13}
         & \multicolumn{3}{c}{$\lambda=10$k} & \multicolumn{3}{c}{$\lambda=100$k} 
         & \multicolumn{3}{c}{$\lambda=10$k} & \multicolumn{3}{c}{$\lambda=20$k} 
         \\ 
         \cmidrule(lr){2-4} \cmidrule(lr){5-7} \cmidrule(lr){8-10} \cmidrule(lr){11-13}
         & ppl & ppl@sw & rec & ppl & ppl@sw  & rec
         & ppl & ppl@sw & rec & ppl & ppl@sw  & rec\\ \hline
         \multicolumn{1}{l}{Ind. LSTM} & $7.1$ & $7.16$ & $1.15$ & $4.7$ & $4.73$ & $1.18$ & $356$ & $349$ & $1.11$ & $295$ & $292$ & $1.15$ \\\hline
         \multicolumn{1}{l}{Large LSTM} & $7.78$ & $10.4$ & $6.82$ & $\textbf{4.86}$ & $8.58$ & $18.9$ & $352$ & $406$ & $3.61$ & $457$ & $619$ & $6.56$ \\
       \multicolumn{1}{l}{MoS} & $8.13$ & $10.6$ & $6.6$ & $5.43$ & $10.3$ & $19$ & $343$ & $443$ & $4.6$ & $298$ & $409$ & $6.08$ \\ 
        \multicolumn{1}{l}{Ensemble 5} &
        $8.84$ & $11.3$ & $7.41$ & $5.6$ & $10.2$ & $24.7$ & $418$ & $519$ & $3.89$ & $317$ & $411$ & $4.83$ \\
       \multicolumn{1}{l}{MoE 5} & 
        $8.65$ & $10.9$ & $7.11$ & $5.55$ & $9.86$ & $24$ & $425$ & $524$ & $3.76$ & $335$ & $439$ & $4.94$ \\
        \multicolumn{1}{l}{MoE+PW 5} &
         $8.74$ & $11.1$ & $7.2$ & $5.58$ & $10$ & $23.3$ & $446$ & $557$ & $3.94$ & $331$ & $432$ & $4.63$ \\
         \multicolumn{1}{l}{PoE 5} & $7.68$ & $10.1$ & $7.06$ & $5.32$ & $9.79$ & $25.5$& $297$ & $389$ & $5.18$ & $404$ & $505$ & $4.47$ \\
         \multicolumn{1}{l}{PoE+PW 5} & $\textbf{7.2}$ & $\textbf{8.46}$ & $\textbf{3.67}$ & ${5.02}$ & ${7.54}$ & ${14.9}$ & $320$ & $361$ & $2.82$ & $270$ & $322$ & $\textbf{3.35}$ \\
        \multicolumn{1}{l}{Ensemble 30} &
        $11.9$ & $14.8$ & $8.08$ & $7.05$ & $14.2$ & $30.9$ & $509$ & $623$ & $3.72$ & $391$ & $511$ & $5.14$ \\
       \multicolumn{1}{l}{MoE 30} &
        $11.1$ & $13.7$ & $7.54$ & $6.89$ & $13.7$ & $30$ & $539$ & $651$ & $3.47$ & $436$ & $572$ & $4.97$ \\
        \multicolumn{1}{l}{MoE+PW 30} &
        $11.2$ & $13.8$ & $7.97$ & $6.92$ & $13.7$ & $29.7$ & $555$ & $675$ & $3.49$ & $419$ & $561$ & $5.43$ \\
         \multicolumn{1}{l}{PoE 30} & $7.96$ & $10.7$ & $7.33$ & $5.17$ & $9.9$ & $24.8$ & $315$ & $375$ & $3.89$ & $297$ & $389$ & $5.18$ \\
         \multicolumn{1}{l}{PoE+PW 30} & $7.41$ & $9.17$ & $4.76$ & $5.04$ & $\textbf{7}$ & $\textbf{9.03}$ & $\textbf{285}$ & $\textbf{316}$ & $\textbf{2.68}$ & $\textbf{241}$ & $\textbf{287}$ & $3.54$ \\
         \hline
        \end{tabular}
    \end{small}
    \caption{Average perplexity (ppl), perplexity for 10 batches after a switch (ppl@sw), and recovery time after a switch in batches (rec) for both datasets per mean sequence length ($\lambda$).}
    \label{tab:results}
\end{table*}
Results are averaged over ten different runs and reported in Table
\ref{tab:results}. Standard deviations are reported on the Supplementary 
Materials.

We begin by observing that higher values of $\lambda$ correspond to lower
perplexities, as expected from the fact that these corpora with longer sequence
lengths are also proportionally larger in total length.

Second, we note that Ensemble and MoE systems with 5 modules and larger hidden
vectors outperformed models with 30 modules and smaller hidden dimensionality,
but this is not the case for PoE, which show comparable performance between 
the two variants or even the opposite trend. 
Furthermore, the PoE's performance is considerably better than the former
two, which can be attributed to a combination of multiple factors. 
On the one hand, we note that, in agreement with previous work~\citep{ShenOAR19},
MoE models are often sensitive to a Winner-Takes-All (WTA) effect in which only one
single expert gets trained at the end.
Thus, models with larger dimensionality per module can benefit from having 
a larger capacity.
However, also ensembles show comparable performance, showing that this effect
is not only caused by a single module being trained.
Perhaps, more important is the fact that, as hypothesized by
\citet{hinton1999products}, PoE can use their capacity to learn complementary
parts of the distribution, and thus it makes a smaller difference for them
whether there are a few high-capacity modules or many of them, but with smaller
capacity.

\begin{figure*}[t]
   \centering
    \begin{subfigure}[b]{0.3\linewidth}
    \centering
         \includegraphics[width=.9\linewidth]{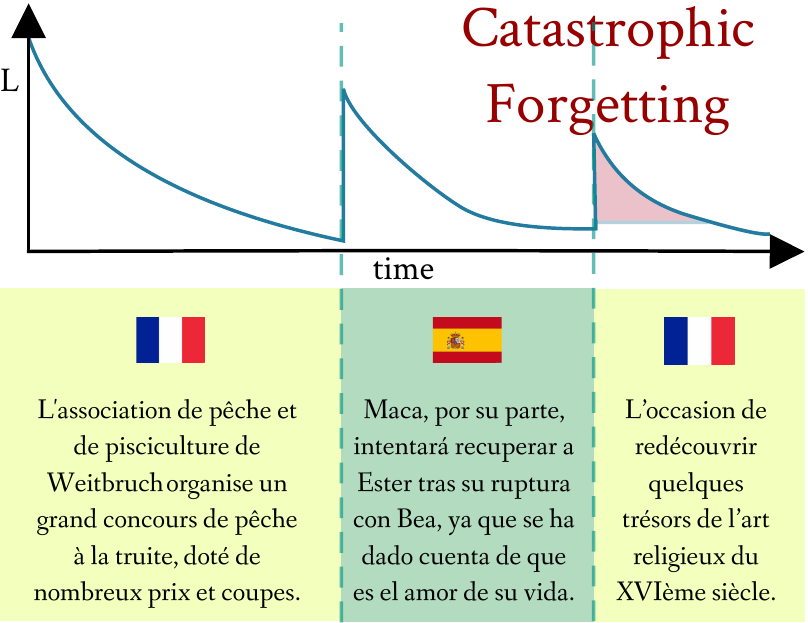}
         \caption{}
    \label{fig:task}
    \end{subfigure}%
    \begin{subfigure}[b]{0.3\linewidth}
    \centering
    \includegraphics[width=.9\linewidth]{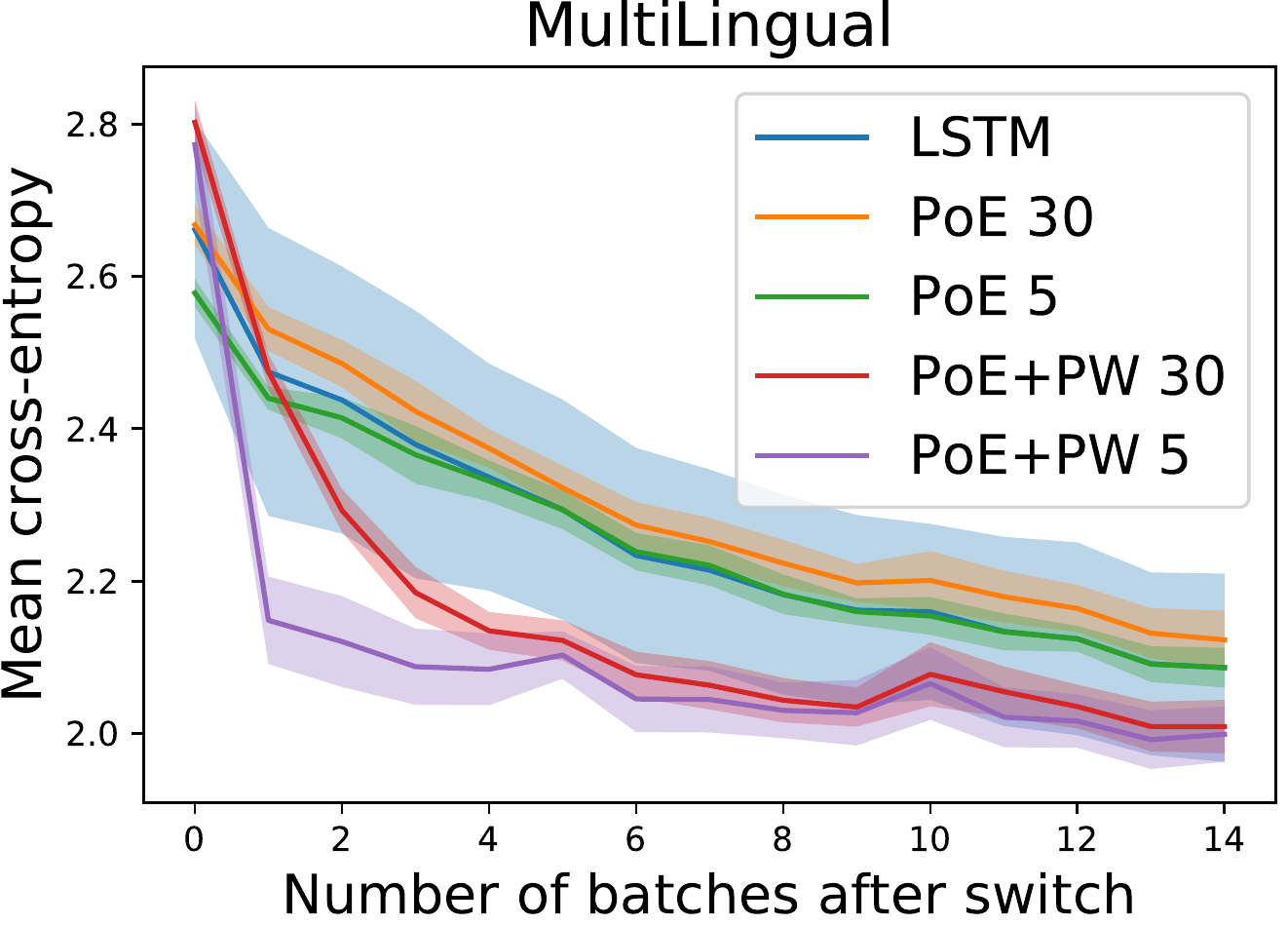}
    \caption{}
    \label{fig:news-switch}
    \end{subfigure}%
    \begin{subfigure}[b]{0.3\linewidth}
    \centering
    \includegraphics[width=.9\linewidth]{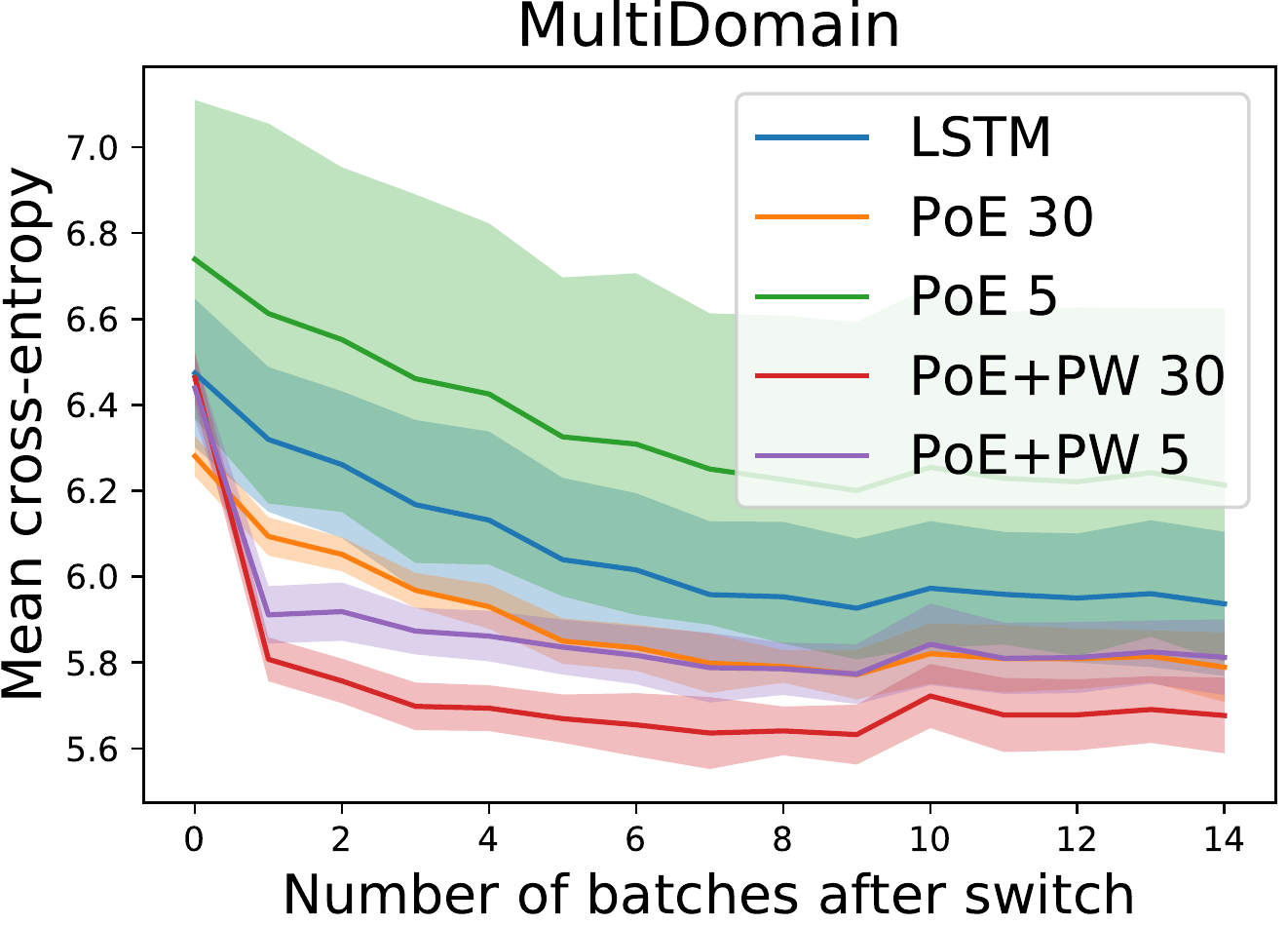}
    \caption{}
    \label{fig:domain-switch}
    \end{subfigure}
    \begin{subfigure}[c]{0.3\linewidth}
        \centering
        \includegraphics[width=\linewidth]{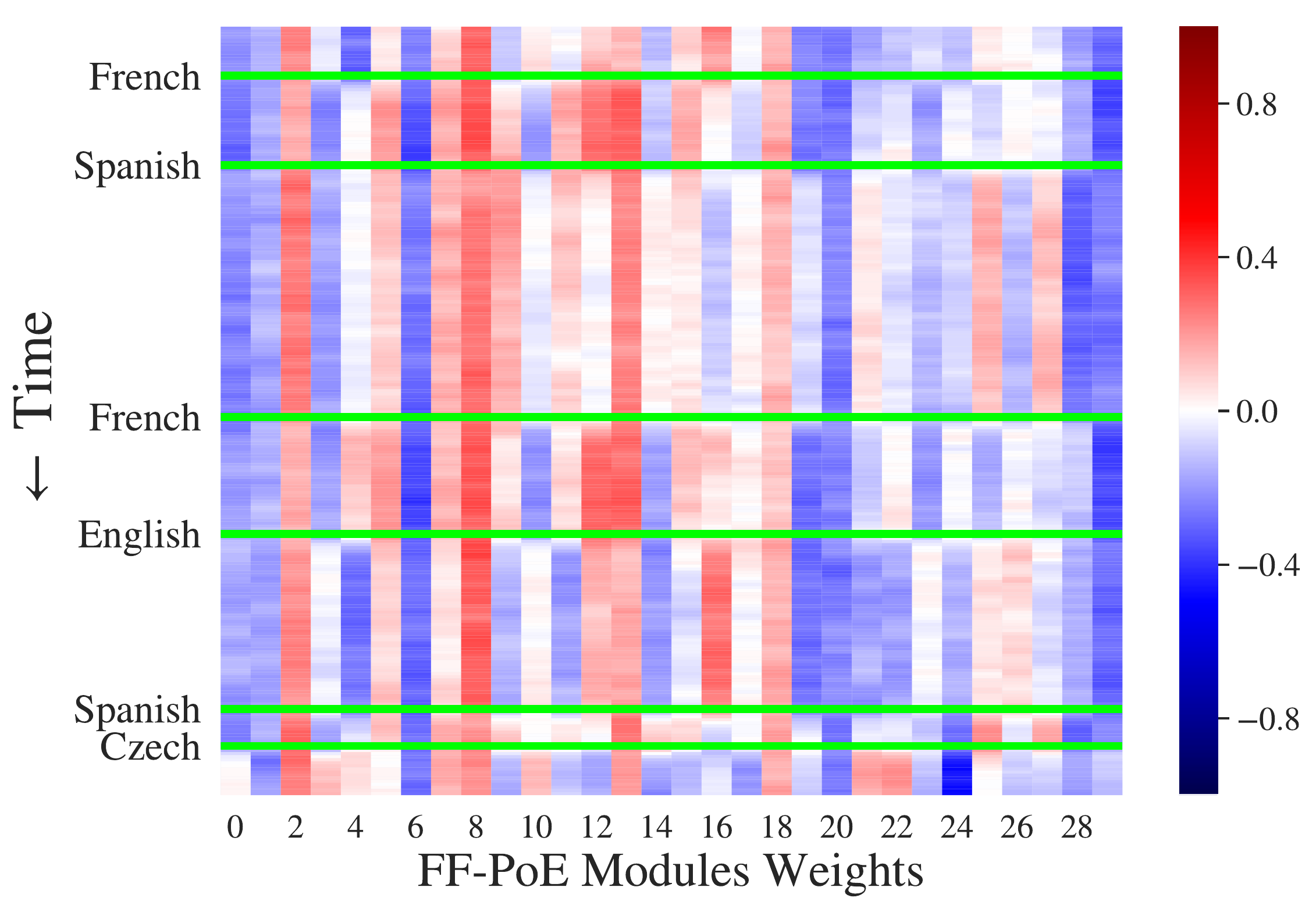}
        \caption{}
        \label{fig:weights}
    \end{subfigure}
    \begin{subfigure}[c]{0.3\linewidth}
        \centering
        \includegraphics[width=0.9\linewidth]{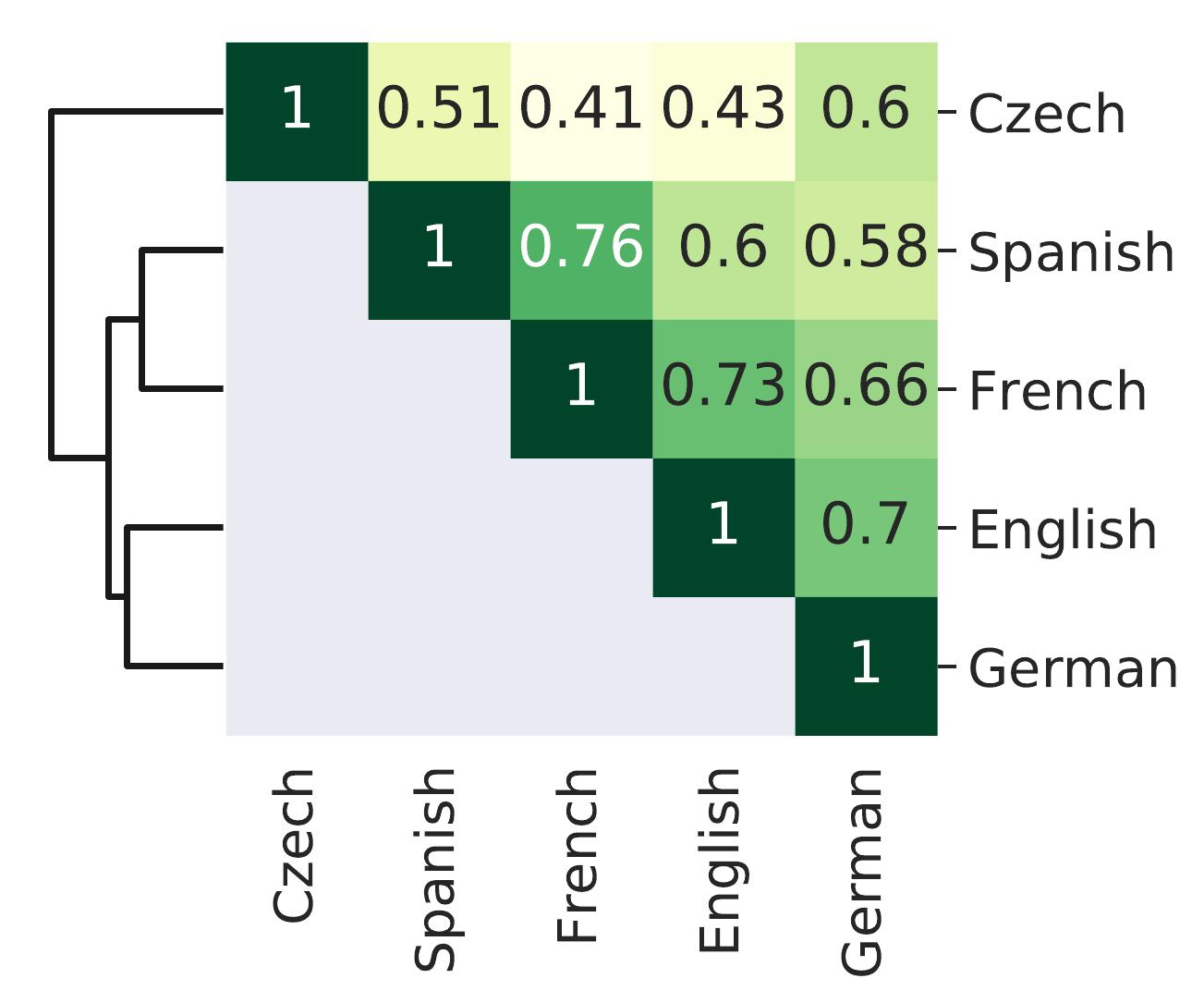}
        \caption{}
        \label{fig:correlations-lang}
    \end{subfigure}
    \caption{(a) CALM: A model's loss ($L$)
         is tracked as it observes text pertaining to different classes, while catastrophic
         interference provokes spikes in this signal. (b-c) Mean cross-entropy for the first 15 batches after a switch averaged over all occurrences in MultiLingual and MultiDomain, respectively, under different random seeds. (d) Mixture weights produced by the PoE+PW 30  model on multilingual data ($\lambda=10$k). (e) Correlation coefficients between mixture weights corresponding to different languages for the PoE+PW 30 model collected during the last 100 batches ($\lambda=10$k).}
\end{figure*}

Next, we note that while PW does not strongly alter performance on MoE
architectures, as expected from the WTA effect influencing these models, they
significantly improve the vanilla PoE counterparts, confirming the
effectiveness of the proposed mechanism in this task.
This observation holds not only for overall perplexity but also in terms of the
metrics quantifying adaptation efficiency at class switches (ppl@sw and rec).
Indeed, Figures \ref{fig:news-switch} and \ref{fig:domain-switch} show this
fact in more detail, by representing the mean cross-entropy of each different
model for the 15 batches occurring immediately after a switch.
As we can see, the PoE+PW model shows a large spike on the first batch because
its adaptation mechanism that depends on this error signal has not kicked in
yet.
However, in the subsequent batch, its performance increases sharply outperforming
comparable models.

In comparison to a monolithic LSTM model, PoE and PoE+PW models perform on-par
on MultiLingual (although with better adaptation records), and better on MultiDomain.
In the latter case, we can observe that the version with 30 modules yields
better performance than the one with just 5.
One possible explanation is related to word-level language modelling being a
higher rank problem than character-level language modelling, and thus it can be
better fitted by combining the judgements from multiple lower-rank
experts~\citep{Yang:etal:2018}.
This explanation is also consistent with the comparatively better performance
of the MoS model.

Finally, we note that the model with Independent LSTMs for each class
performs best on MultiLingual, but it is outperformed by a large margin on
MultiDomain.
We note that this model does not suffer from forgetting when switching classes
but also misses the training signal from transferable training data.
As a consequence, it has an edge on MultiLingual, which switches between
classes that have considerably different statistical properties, but not on
MultiDomain where the differences between the classes are much more nuanced. 
All in all, this shows that while, in consonance with previous
results~\citep{dhar2018does}, there is little room for transferring knowledge
in the MultiLingual case, the MultiDomain setting provides plenty of
opportunities for transferring knowledge across each domain.
Thus domain-agnostic systems can benefit from them.

\subsection{Analysis}
\label{sec:analysis}


Next, we turned to analyze the gating strategies acquired by the more successful
models to understand whether they have captured the latent similarities 
between classes and how might they help them in coping with catastrophic
forgetting.
For this, we focused on the PoE+PW 30 modules operating on the MultiLingual
dataset ($\lambda=10$k) because its 30-dimensional gate vectors can represent
more nuanced similarities.

Figure \ref{fig:weights} shows a heatmap of the mixture weights as the model processes
different language sequences.
High absolute values represent the activation of a module, regardless of
whether these are negative or positive values.
It can be seen that upon language switches, the model reconfigures itself to a
different set of mixture weights that are maintained more or less consistently
within the sequence.
Furthermore, we note that modules that receive mixture weights close to $0$
are protected from forgetting, as this gating value is also multiplied to the
module's gradients.
Moreover, we hypothesize that modules are protected even when their 
corresponding weight is set to the opposite sign (see, for instance, module 16
on English and Spanish), because the incoming training data serves as negative
training data, namely, something not-to-be-predicted. 
Thus, this should not affect what the model does predict when used with a 
positive weight.
Instead, this allows for dual use of the modules, encoding information both
when it is weighted positively and negatively.

Finally, recall from Section \ref{sec:dataset} that the languages in our
MultiLingual dataset are derived from different linguistic families with a
latent similarity structure.
To uncover whether the learned latent similarities reflect this structure, we
computed the correlations between the mixture weights produced while processing
the last 100 batches of each class.
The results are displayed in Figure \ref{fig:correlations-lang}, and show 
that the similarities are indeed well-reflected in the gating values.
Notably, we observe that Czech seems to be using the most distinct set of
modules. 
Spanish and French correlate quite strongly in the modules they use, and
while English also correlates with French, it also does so with German, with
the latter correlating to a lesser extent with the other languages.
Indeed, applying a simple hierarchical clustering algorithm over this matrix
recovers the underlying linguistic families!


\section{Conclusions}
\label{sec:conclusions}

In this paper, we have introduced the class-agnostic continual language
modelling task (CALM), together with a Python library with MultiLingual and
MultiDomain datasets, which allows multiple parameter configurations and can
also be easily adapted to different corpora.
We expect that it will foster more empirical work on continual learning in a
language-centred setup in which there is a natural latent similarity structure
between different tasks.
We have argued that in addition to measuring the overall performance of models
in an online learning fashion, their susceptibility to catastrophic forgetting
can be observed in terms of adaptation speed to changes in the input class, and
proposed measures to capture it.
Finally, we have evaluated multiple simple baselines to serve as references for
future work on this benchmark and introduced a simplification of the gating
strategy for a Product of Experts, which improves its performance significantly 
by allowing it to distribute effectively different distributions across
different experts so that the resulting system can act as a strong baseline for
future work on this task.

While addressing catastrophic forgetting is still a major challenge for Online
Continual Learning, it is by no means the only one. 
In the future, we would like to understand how learning systems can also
bootstrap on their knowledge to improve their learning skills, so that they
will not only be able to acquire knowledge from different sources in a seamless
way but also get better at it as they go.

\subsubsection*{Acknowledgments}

We would like to thank Marco Baroni, Marc'Aurelio Ranzato, and the numerous
anonymous reviewers that have provided valuable comments on earlier stages of
this work.
%

\bibliography{bibliography}
\bibliographystyle{iclr2021/iclr2021_conference.bst}

\appendix
\label{appendix}
\section{Corpus examples}
Figure \ref{fig:multilingual-samples} and \ref{fig:multidomain-samples} present samples from the corpora used for our dataset. As stated in the paper, we can notice a much bigger difference between input class in the case of the multilingual setup, while the differences in the case of the multidomain setup are more subtle and nuanced.

\begin{figure}[ht]
\small
\begin{tabular}{c|c}
& Dataset samples \\ \hline
czech & \begin{tabular}[c]{@{}c@{}}Maďarská iNFiNITY Coliseum Lan je pokračováním \\ úspěšného BECUPu, z něhož  si nejeden náš tým v minulosti odvezl medaili.\end{tabular} \\ \hline
english & \begin{tabular}[c]{@{}c@{}}If Hofmann played the role of paterfamilias, Anaïs Nin was the\\ bad mother to Admiral  and De Niro's group. This one wasn't close.\end{tabular} \\ \hline
french & \begin{tabular}[c]{@{}c@{}}Le Beatle s'en est alors emparé pour  créer un chef-d'oeuvre psychédélique\\  longtemps associé à l'usage du LSD.\end{tabular}  \\\hline
german & \begin{tabular}[c]{@{}c@{}}Im ersten Jahr hatten sie schon 278  Anfragen, fast\\ 60 ehemalige Manager  und Unternehmer wollten mitmachen.\end{tabular} \\ \hline
spanish & \begin{tabular}[c]{@{}c@{}}Los despidos serán realizados por medio del plan de GM de cese de empleo, por \\ lo que no se ofrecerán jubilaciones anticipadas \end{tabular} 
\end{tabular}
\caption{Samples from the multilingual dataset}
\label{fig:multilingual-samples}
\end{figure}

\begin{figure}[ht]
\small
\begin{tabular}{c|c}
 & Multidomain dataset samples \\ \hline
bnc & \begin{tabular}[c]{@{}c@{}}Good weather for the crops. Have your sheep been suffering much from the staggers ? \\ Have you contributed a great deal  this year to the butter mountain ?\end{tabular} \\ \hline
euro & \begin{tabular}[c]{@{}c@{}}I would like your advice about Rule 143 concerning  inadmissibility. \\ My question relates  to something that will come up on Thursday\end{tabular} \\ \hline
news & \begin{tabular}[c]{@{}c@{}}If Hofmann played the role of paterfamilias, Anaïs Nin was the\\ bad mother to Admiral  and De Niro's group. This one wasn't close.\end{tabular} \\ \hline
wiki & \begin{tabular}[c]{@{}c@{}}Otto , Prince of Bavaria , was chosen as the first  King of Greece in 1832 , under the name Othon .\\  His arrival in Nafplio , then the Greek capital, was hailed enthusiastically by Makriyannis\end{tabular}
\end{tabular}
\caption{Samples from the multi-domain dataset}
\label{fig:multidomain-samples}
\end{figure}
\section{Further analysis}
\subsection{PoE weights behaviour}

We also inspected the gate values produced by LSTM-gated PoE models observing
that the models are indeed not learning a class-switching mechanism.
We hypothesized that this is due to the fact that when the experts are still untrained, the LSTM produces some arbitrary but consistent gating values, making those selected modules being the only ones to be trained, and thus falling into a vicious cycle.
As a sanity check that supports this hypothesis, we first pre-trained a set of modules while still using our simple gating mechanism. Then, we initialized with these modules a network that now used LSTM mixture weights, but training on very short sequences to avoid the effect of catastrophic forgetting affecting the network. In this context, the network learned the appropriate gating as expected.

\subsection{MultiDomain module correlation}
In comparison with the Multilingual setup, correlations in the MultiDomain case are much weaker. Moreover, they are weak even within the same class: When we measure the
autocorrelation between weights pertaining to the last 100 batches with the
preceding 100 ones we obtain values in the order of $0.65$, much lower than for
the MultiLingual experiments, where they are in the order of $0.96$ (see Figure \ref{fig:correlations-domain}).
This shows that model usage is less consistent per-class, which could be
explained by the fact that classes are much more nuanced than in MultiLingual
and their corresponding distributions are far more complex.
These results are also consistent with our experimental observation that the
MultiDomain dataset was more amenable to transfer between different classes
than the MultiLingual, as these classes could be distributed more evenly across
the model and could be characterized with multiple mixture weights
configurations.

\begin{figure*}[h]
    \centering
      \begin{subfigure}[b]{0.35\linewidth}
        \centering
        \includegraphics[width=\linewidth]{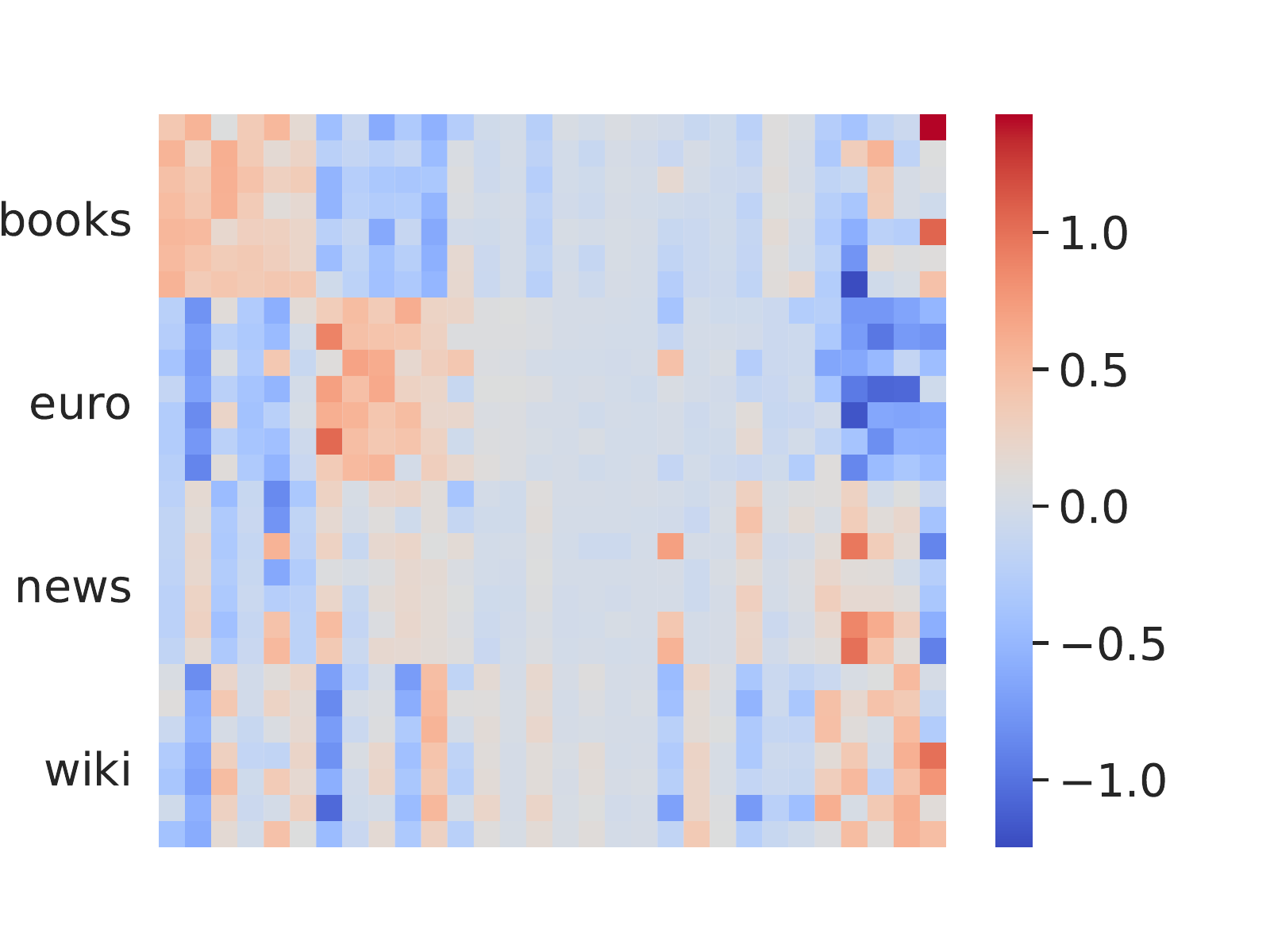}
\caption{Heatmap of weights with rows sorted by domain and columns sorted by similarity}
        \label{fig:correlations-domain}
    \end{subfigure}
      \begin{subfigure}[b]{0.35\linewidth}
        \centering
        \includegraphics[width=.9\linewidth]{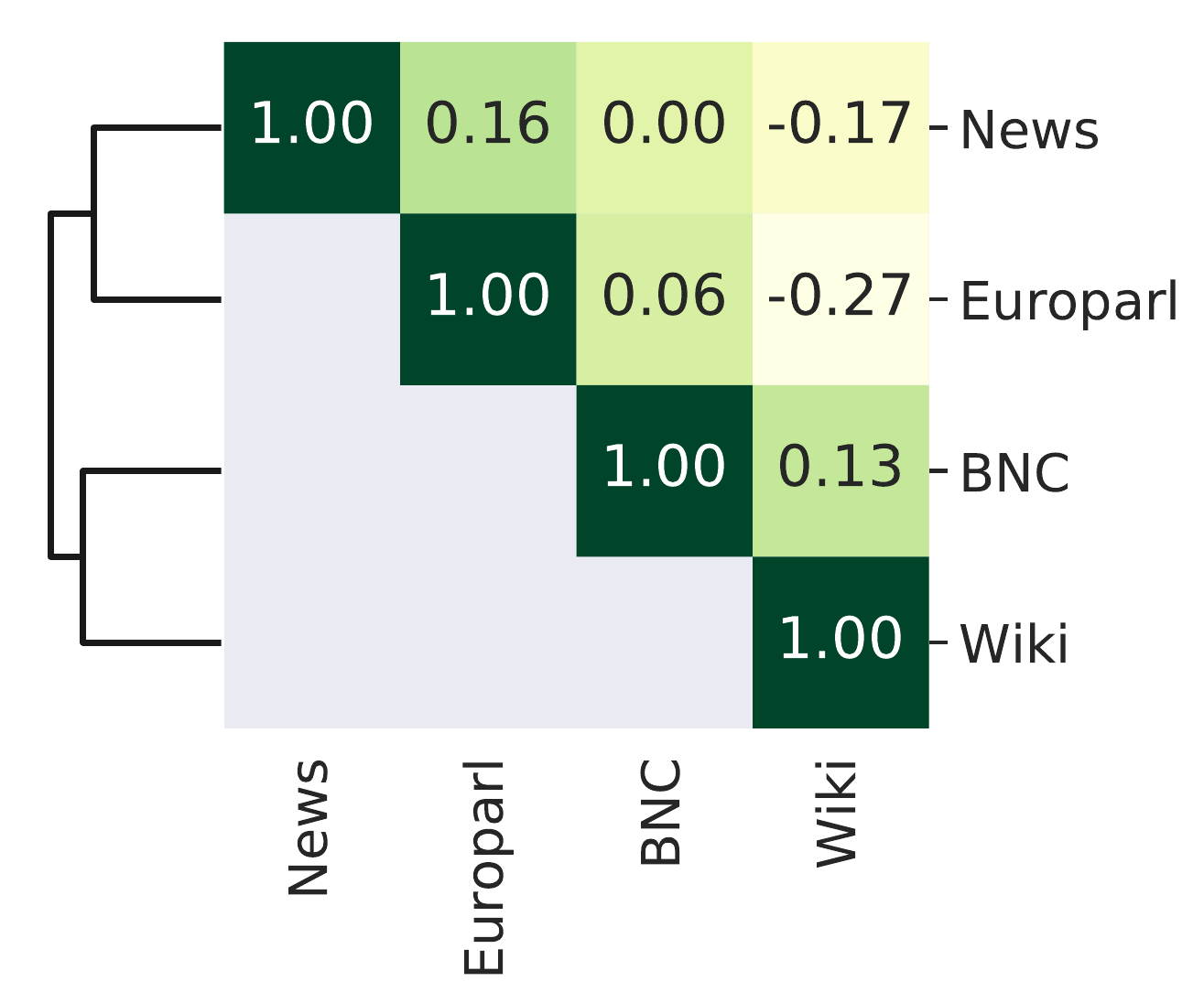}
        \caption{Weight correlations}
        \label{fig:correlations-domain}
    \end{subfigure}
    \caption{MultiDomain ($\lambda$=10k) analysis}
\end{figure*}

\section{Transformer Experiments}

We experimented extensively with Transformer models. 
One difference with respect to LSTM models is that Trasnformers, at least in
their vanilla versions, are not autoregressive, and thus they cannot transfer
information from the past. 
In standard NLP tasks, they largely overcome this problem by using a large
context window on which they can operate effectively.
Thus, to afford them similar memory capabilities, we kept a buffer of the last
$b\times512$ consecutive examples that was continually updated with each
incoming mini-batch.

\citet{vaswani2017attention} describes a learning rate scheduling scheme 
in which the learning rate is linearly increased until reaching a number of 
warmup steps, and then annealed from that point on.
Considering that in a Continual Learning setup the model is not expected
to converge, annealing might not be appropriate. 
Thus, we also experimented with keeping the learning rate flat after 
that point on.
We experimented with both learning rate schedules, plus no scheduling at all.
Furthermore, we considered both training with default Adam parameters
($\alpha=10^{-3}, \beta=(0.9, 0.999)$) or the ones reported by \citet{vaswani2017attention}
and base learning rates of $\frac{1}{d_{model}}$, $1e^{-3}$, $5e^{-3}$, $0.01$.
Also, we tuned the warmup steps among $400$, $2000$ and $4000$ steps.
The best perplexity results we obtained in the Multilingual validation data
were $13.2$ for $\lambda=10$k and $6.02$ for $\lambda=100$k, whereas in the 
Multidomain validation data we got $686$ average perplexity for $\lambda=10$k
and $527$ for $\lambda=20$k.

\section{Generated output}
In Figure \ref{fig:output-ffpoe}, we present generated samples from different stages of training. 
These generated examples are produced by sampling one character at a time from the models, and using them as input for the next time step.
As quantitatively observed in the paper, it adapts much faster to the current input type (French) in comparison with an LSTM, which generates text resembling the language of the previously seen class even after 10 batches.

\begin{figure}[ht]
\small
\centering
\captionsetup{justification=centering}
\begin{tabular}{c|c|c}
 & PoE+PW 30 & LSTM\\ \hline
\begin{tabular}[c]{@{}c@{}}end\\ english\end{tabular} & \begin{tabular}[c]{@{}c@{}}for a release was the week in Troust \\ Pglates in George Services are\\  claimed whet this could get one\end{tabular} & \begin{tabular}[c]{@{}c@{}}Tvice. (Relátórs had the state's\\  annual annual Call Statua plannting\\  more years' physical cost\end{tabular}\\ \hline
\begin{tabular}[c]{@{}c@{}}5 batches\\ french\end{tabular} & \begin{tabular}[c]{@{}c@{}}lement Filmarian Roads. Aus cadres\\  et temps disputer Lileana Maan. \\ Institution, le provinces, unbieut\end{tabular} & \begin{tabular}[c]{@{}c@{}}eau polítical but the room of Noxe\\  Common Electrical Taladei Baritef.\\ BAG - Runey premium begai maki\end{tabular}\\ \hline
\begin{tabular}[c]{@{}c@{}}10 batches\\ french\end{tabular} & \begin{tabular}[c]{@{}c@{}}Definit werde à l'équipe pass,\\  libertant Youth Losier Chavez and\\  Jean-Pierro. «Vu entre fascal publ\end{tabular} & \begin{tabular}[c]{@{}c@{}}attempted Jueves Mo., unit encome\\  ergarded a next post television \\ genetical dangere tet. For hemous\end{tabular}\\ \hline
\begin{tabular}[c]{@{}c@{}}end \\ french\end{tabular} & \begin{tabular}[c]{@{}c@{}}ive commune services au \\ championnat où qui se sont renfovées\\  de la hierre du 23,4er est dit doubles\end{tabular} & \begin{tabular}[c]{@{}c@{}}el-Bilanze extranger à la fin de\\  l‘Etat: "Yens ni irneu à Show Joban\\  ? Il vio, les grandes hommes de\end{tabular}
\end{tabular}
\caption{Generated text at different stages\\ of training}
\label{fig:output-ffpoe}
\end{figure}

\section{Model sizes}
As it is shown in Table \ref{tab:hidden-size}, the number of hidden units varies for most of the models. We vary the hidden size in order to keep a similar number of parameters across the models: around 22 million for the multilingual setup and around 600 million for the multidomain one.

\begin{table}[h]
    \centering
    \begin{small}
      \begin{tabular}{ccccc}
        \cline{2-5}
          & \multicolumn{2}{c}{MultiLingual} & \multicolumn{2}{c}{MultiDomain}\\         
         \cmidrule(lr){2-3} \cmidrule(lr){4-5}
        Model & Hidden size & \#Parameters & Hidden size & \#Parameters \\\hline
        LSTM & $1300$ & $21.66$M & $5200$ & $605.2$M \\
        Ind. LSTM & $550$ & $20.2$M & $1800$ & $571.2$M \\
        PoE/MoE (+PW) 5 & $550$ & $21.2$M & $1600$ & $621.8$M \\ 
        PoE/MoE (+PW) 30 & $200$ & $21.85$M & $200$ & $635.3$M \\ 
        MoS & $500$ & $22$M & $2620$ & $572$M\\
    \end{tabular}
    \end{small}
    \caption{Model sizes}
    \label{tab:hidden-size}
\end{table}

\section{Hyperparameter search} \label{hyperparameter}

\begin{table*}[ht]
\small
\centering
\begin{tabular}{|c|c|c|c|c|c|c|c|c|c|}
\hline
task & $\lambda$ & model & nhid & dropout & \begin{tabular}[c]{@{}c@{}}learn\\ iter.\end{tabular} & \begin{tabular}[c]{@{}c@{}}adapt.\\ iter.\end{tabular} & modules & \begin{tabular}[c]{@{}c@{}}gating\\ nhid\end{tabular} & \begin{tabular}[c]{@{}c@{}}clear\\ gating\end{tabular} \\ \hline
\multirow{6}{*}{lang.} & \multirow{3}{*}{10k} & lstm & 200, \textbf{1300} & \begin{tabular}[c]{@{}c@{}}0.1, \textbf{0.2},\\ 0.4\end{tabular} & 1, \textbf{2}, 5 & - & - & - & - \\ \cline{3-10} 

 &  & MoE/PoE & \textbf{200}, \textit{550} & \textbf{\textit{0.2}} & \textbf{\textit{2}}, 5 & \textbf{\textit{1}} & \textit{5}, \textbf{30} & \begin{tabular}[c]{@{}c@{}} 50, \textit{100}\\ \textbf{200}\end{tabular} & \textbf{0}, \textit{1} \\ \cline{3-10} 

 &  & MoE/PoE+PW & \textbf{200}, \textit{550} & \textbf{\textit{0.2}} & \textbf{\textit{2}}, 5 & 1, \textbf{10}, \textit{100} & \textit{5}, \textbf{30} & - & - \\ \cline{2-10} 
 & \multirow{3}{*}{100k} & lstm & 200, \textbf{1300} & \begin{tabular}[c]{@{}c@{}}\textbf{0.1}, 0.2,\\ 0.4\end{tabular} & \textbf{1}, 2, 5 & - & - & - & - \\ \cline{3-10}
  
 &  & MoE/PoE & \textbf{200}, \textit{550} & \textbf{\textit{0.2}} & \textbf{\textit{1}}, 2, 5 & \textbf{\textit{1}} & \textit{5}, \textbf{30} & \begin{tabular}[c]{@{}c@{}}50, 100\\ \textbf{\textit{200}}\end{tabular} & 0, \textbf{\textit{1}} \\ \cline{3-10} 

 &  & MoE/PoE+PW & \textbf{200}, \textit{550} & \textbf{\textit{0.2}} & \textbf{\textit{1}}, 2, 5 & 1, 10, \textbf{\textit{100}} & \textit{5}, \textbf{30} & - & - \\ \hline
\multirow{6}{*}{dom.} & \multirow{3}{*}{10k} & lstm & \textbf{5200} & \begin{tabular}[c]{@{}c@{}}\textbf{0.1}, 0.2,\\ 0.4\end{tabular} & 1, 2, \textbf{5} & - & - & - & - \\ \cline{3-10} 

 &  & MoE/PoE & \textbf{200}, \textit{1600} & \textbf{\textit{0.2}} & 1, \textbf{\textit{2}}, 5 & \textbf{\textit{1}} & \textit{5}, \textbf{30} & \begin{tabular}[c]{@{}c@{}}\textit{50}, 100\\ \textbf{200}\end{tabular} & \textit{0}, \textbf{1} \\ \cline{3-10} 

 &  & MoE/PoE+PW & \textbf{200}, \textit{1600} & \textbf{\textit{0.2}} & \textit{1}, \textbf{2}, 5 & 1, 10, \textbf{\textit{100}} & \textit{5}, \textbf{30} & - & - \\ \cline{2-10} 

 & \multirow{3}{*}{20k} & lstm & \begin{tabular}[c]{@{}c@{}}200, 1300, \\ \textbf{5200}\end{tabular} & \begin{tabular}[c]{@{}c@{}}0.1, 0.2,\\ \textbf{0.4}\end{tabular} & \textbf{1}, 2, 5 & - & - & - & - \\ \cline{3-10} 

 &  & MoE/PoE & \textbf{200}, \textit{1600} & \textbf{\textit{0.2}} & \textit{1}, 2, 5 & \textbf{\textit{1}} & \textit{5}, \textbf{30} &\begin{tabular}[c]{@{}c@{}} 50, 100\\ \textbf{\textit{200}}\end{tabular} & \textbf{\textit{0}},1 \\ \cline{3-10} 

 &  & MoE/PoE+PW & \textbf{200}, \textit{1600} & \textbf{\textit{0.2}} & \textbf{\textit{1}}, 2, 5 & 1, 10, \textbf{\textit{100}} & \textit{5}, \textbf{30} & - & - \\ \hline
\end{tabular}
\caption{Table with the hyperparameters tested on the models: LSTM, PoE, and PoE+PW. The bold parameters are the ones chosen for LSTM, MoE/PoE-30, MoE/PoE+PW 30 and the italic parameters are the ones chosen for MoE/PoE-5 and MoE/PoE+PW 5}
\label{table:hyper-models}
\end{table*}

Table \ref{table:hyper-models} present the explored hyperparameters for LSTM and PoE. The parameters in bold are the ones chosen for the final models, with the exception of PoE-5 and PoE+PW 5 which are marked with italics. 

The meaning of the different hyperparameters for Table \ref{table:hyper-models} is:
\begin{itemize}
\item nhid: the size of the hidden state of the base LSTM
\item dropout: the dropout value used in the base module of the LSTM
\item learn iter.: how many learning iterations over each batch are done before moving to the next batch
\item adapt. iter.: it is used in the case of PoE+PW and it shows how many iterations to train the gating weights are done for each learning iteration.
\item modules: how many modules does the PoE models contain
\item gating nhid: the size of the hidden state for the LSTM used to calculate the gating weights in the case of PoE
\item clear gating: it is a boolean value which clears the hidden state of the LSTM used for gating weights in the case of PoE
\end{itemize}


Also, MoS was tuned following the hyperparameters: 1 or 2 learning iterations, learning rate 1e-3 or 5e-4. For the domain setup, we considered the combinations: (nsoftmaxes=2, nhid=4750) or (nsoftmaxes=50, nhid=2620). On the other hand, for the multilingual dataset, we considered (nsoftmaxes=2, nhid=1200) or (nsoftmaxes=75, nhid=500).

\section{Standard Deviations}
\begin{table*}[ht]
    \setlength{\tabcolsep}{2.5pt}
        \begin{small}
         \centering
           \begin{tabular}{ccccccccccccc}
        \cline{2-13}
         & \multicolumn{6}{c}{MultiLingual} & \multicolumn{6}{c}{MultiDomain}\\         
         \cmidrule(lr){2-7} \cmidrule(lr){8-13}
         & \multicolumn{3}{c}{$\lambda=$ 10k} & \multicolumn{3}{c}{$\lambda=$ 100k} 
         & \multicolumn{3}{c}{$\lambda=$ 10k} & \multicolumn{3}{c}{$\lambda=$ 20k} 
         \\ 
         \cmidrule(lr){2-4} \cmidrule(lr){5-7} \cmidrule(lr){8-10} \cmidrule(lr){11-13}
         & ppl & ppl@sw & rec & ppl & ppl@sw  & rec
         & ppl & ppl@sw & rec & ppl & ppl@sw  & rec\\ \hline
         \multicolumn{1}{l}{Ind. LSTM} & $0.42$ & $0.41$ & $0.44$ & $0.12$ & $0.05$ & $0.3$
         & $28.5$ & $25.5$ & $0.2$ & $17.1$ & $16.6$ & $0.22$ \\\hline
         \multicolumn{1}{l}{Large LSTM} & $1.08$ & $1.81$ & $0.98$ & $0.28$ & $0.87$ & $2.84$
         & $51$ & $98.1$ & $0.64$ & $-$ & $-$ & $-$ \\
        \multicolumn{1}{l}{MoS} & $0.5$ & $0.8$ & $0.87$ & $0.15$ & $0.55$ & $1.82$ & $32.4$ & $40.7$ & $0.2$ & $19.9$ & $18.5$ & $0.4$ \\
         \multicolumn{1}{l}{PoE 5} & $0.23$ & $0.2$ & $0.7$ & $0.14$ & $0.24$ & $2.51$
         & $229$ & $332$ & $0.32$ & $18.8$ & $19.6$ & $0.5$ \\
         \multicolumn{1}{l}{PoE 30} & $0.28$ & $0.28$ & $0.72$ & $0.12$ & $0.21$ & $1.5$
         & $27.7$ & $14.4$ & $0.22$ & $15.4$ & $15.1$ & $0.48$ \\
         \multicolumn{1}{l}{PoE+PW 5} & $0.17$ & $0.33$ & $0.9$ & $0.11$ & $0.48$ & $3.2$
         & $26.4$ & $22.1$ & $0.52$ & $15.7$ & $19.6$ & $0.62$ \\
         \multicolumn{1}{l}{PoE+PW 30} & $0.21$ & $0.2$ & $0.44$ & $0.1$ & $0.1$ & $0.66$
         & $23.7$ & $16.5$ & $0.3$ & $14.3$ & $12.2$ & $0.28$ \\ 
        \multicolumn{1}{l}{Ensemble 5} &
         $0.249$ & $0.301$ & $0.85$ & $0.12$ & $0.35$ & $2.32$ & $54.5$ & $76.4$ & $0.419$ & $27.6$ & $32.6$ & $0.407$ \\
        \multicolumn{1}{l}{Ensemble 30} &
        $0.375$ & $0.525$ & $0.923$ & $0.203$ & $0.549$ & $2.01$ & $35.1$ & $45.9$ & $0.287$ & $24.5$ & $27.8$ & $0.545$ \\
        \multicolumn{1}{l}{MoE 5} &
         $0.255$ & $0.274$ & $0.855$ & $0.12$ & $0.426$ & $1.75$ & $64.2$ & $84.4$ & $0.447$ & $42.4$ & $56$ & $0.464$ \\
        \multicolumn{1}{l}{MoE 30} &
        $0.21$ & $0.22$ & $0.3$ & $0.1$ & $0.16$ & $2.17$ & $21$ & $23.2$ & $0.53$ & $18$ & $17.2$ & $0.45$ \\  
        \multicolumn{1}{l}{MoE+PW 5} &
        $0.264$ & $0.377$ & $0.806$ & $0.101$ & $0.322$ & $2.36$ & $53.2$ & $71.6$ & $0.36$ & $34.4$ & $40.4$ & $0.445$ \\
        \multicolumn{1}{l}{MoE+PW 30}
        & $0.326$ & $0.561$ & $0.753$ & $0.195$ & $0.543$ & $2.24$ & $43.5$ & $43.8$ & $0.315$ & $26.6$ & $40.3$ & $0.74$ \\
        \hline  
        \end{tabular}
    \end{small}
    \caption{Standard deviation for Average perplexity (ppl), perplexity for 10 batches after a switch (ppl@sw), and recovery time after a switch in batches (rec) for both datasets per mean sequence length ($\lambda$).}
    \label{tab:std_results}

\end{table*}

\end{document}